# Evidential Force Aggregation

Johan Schubert

Department of Data and Information Fusion

Division of Command and Control Systems

Swedish Defence Research Agency

SE-172 90  Stockholm, Sweden

schubert@foi.se

http://www.foi.se/fusion/

**Abstract -** *In this paper we develop an evidential force aggregation method intended for classification of evidential intelligence into recognized force structures. We assume that the intelligence has already been partitioned into clusters and use the classification method individually in each cluster. The classification is based on a measure of fitness between template and fused intelligence that makes it possible to handle intelligence reports with multiple nonspecific and uncertain propositions. With this measure we can aggregate on a level-by-level basis, starting from general intelligence to achieve a complete force structure with recognized units on all hierarchical levels.*

## 1   Introduction

We define force aggregation as a combination of two processes. First, an association of intelligence reports, objects or units (depending on hierarchical level) by a clustering process [1–5] the left column in Figure 1. Secondly, a classification of cluster content through a comparison with templates, the right column in Figure 1. In Figure 1 such clustering and classification is performed on all hierarchical levels, level-by-level to achieve a complete force aggregation of all units.

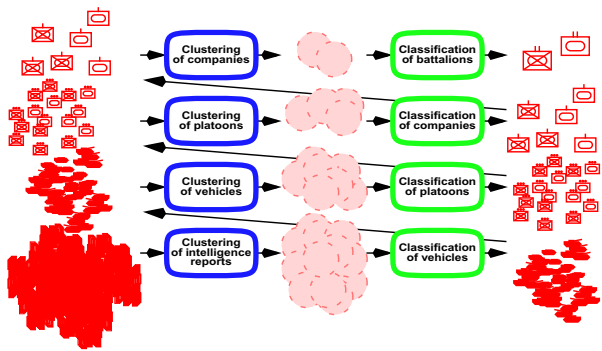

Figure 1: The aggregation process hierarchy.

Evidential force aggregation is force aggregation from uncertain information. The classification in evidential force aggregation is the focus of this paper.

The work described herein is an extension of previous work. In [6] we restricted each intelligence report to carry only one proposition that could be specific or nonspecific regarding object types, i.e., support any subset of all possible types, but was always certain. Here, in this paper, we allow for any number of nonspecific and uncertain propositions in each intelligence report. With this extension we may handle any general intelligence report.

The classification process deals with intelligence reports on a cluster-by-cluster basis. Looking at intelligence in one of the clusters, the classification from intelligence by templates take place in two phases. First, we combine all intelligence reports within the cluster, and secondly, we compare the combined intelligence with all available templates.

In the combination of intelligence a special concern is the representation used. As the reports in general are not reports about the same object or group of objects, we must not use a simple representation dealing only with object type. Instead, we must use a more advanced representation that allow us to keep track of different objects and their possible types. Intelligence reports that actually are referring to the same object or group of objects are precombined, and henceforth viewed as one intelligence report. When this is done, all intelligence reports in the cluster under investigation can be combined, giving us the possibility to investigate the different resulting hypothesis regarding force composition.

When selecting a template for the current cluster we search for a maximum matching between template and fused intelligence. Since intelligence consists of multiple alternative hypothesis with an accompanying uncertainty we must take every hypothesis into account, to its degree of uncertainty, when evaluating a template. As these hypothesis are also nonspecific regarding object type, i.e., they refer to a subset of all possible types instead of to a single type, we cannot expect a perfect matching for each type of object in the template. Instead, we look for possibility of a matching between intelligence and template, i.e., the

1223

absence of conflicts in numbers between what the intelligence propose and what each available template request for all subsets of types. With this measure we can select a template for intelligence with nonspecific propositions.

A few other approaches to force aggregation than the one described here are [7–10].

In Sect. 2 we describe the representation of intelligence and their combination. An example is given in Sect. 2.1. In Sect. 3 we describe the representation of templates and their evaluation and selection through a comparison with intelligence. A continuation of the example is presented in Sect. 3.1. An evidential force aggregation algorithm based on the result of the two previous sections is presented in Sect. 4. Finally, conclusions are drawn (Sect. 5).

## 2 Intelligence

We will here investigate the representation and combination of all intelligence referring to the same unit. We assume that a number of intelligence reports about different set of objects are available. These reports have already been partitioned into subsets where each subset corresponds to a unit on one hierarchical level higher [2, 3]. Let us hereafter focus on one such subset $\chi_a$ and the aggregation of the intelligence in this subset.

Let $TY$ be a set of all possible types of objects $\{TY_x\}$; where $TY_x$ is a type of vehicle or a type of unit depending on which hierarchical level we are at.

Let $I_a$ be a set of any number of intelligence reports in cluster $\chi_a$,

$$I_a = \{C_a^i\}_i. \qquad (1)$$

We use Dempster-Shafer theory [11–16], to represent the uncertainty of all intelligence reports. Each intelligence report focuses on a separate set of objects and is represented by a set of any number of alternative pairs

$$\{(C_a^i \bullet n_j, C_a^i \bullet pt_j)\}_j, \qquad (2)$$

i.e., focal elements. Each pair has a possibly nonspecific proposition about possible types

$$C_a^i \bullet pt_j \subseteq TY \qquad (3)$$

and a corresponding possibly nonspecific number of such types

$$C_a^i \bullet n_j \subset \mathbb{Z}+, \qquad (4)$$

i.e., a subset of $\{1, 2, ...\}$, where

$$N_{C_a} = max_{ij}\{max\ C_a^i \bullet n_j\} \qquad (5)$$

is the maximum number of objects. Each focal element, in the set of Eq. (2), has a basic probability number

$$m_i^{I_a}[(C_a^i \bullet n_j, C_a^i \bullet pt_j)] \qquad (6)$$

indicating the uncertainty in each proposition.

If we receive several reports focused on the same object or set of objects they are precombined into $C_a^i$. Multiple nonalternative propositions about other objects than the $C_a^i \bullet n_j$ objects described as $C_a^i \bullet pt_j$ are handled as additional but separate intelligence reports.

In this situation we have a single frame about the number of objects for each subset of object types $X$,

$$\Theta_X = \{(i, X)\}_{i=0}^{N_{C_a}} \qquad (7)$$

where $X \subseteq TY$.

In order to be able to handle reports about different objects that should not be combined on the object level, but should be viewed as fragments of a larger unit structure where all fragments are to be combined, we need to refine our representation. Each report is now corresponding to a unique position in a unit structure.

The frame of discernment when fusing reports regarding different sets of objects that should be combined as fragments of a larger unit structure becomes

$$\Theta_{I_a} = \{\langle x_1, x_2, ..., x_{|I_a|}\rangle\} \qquad (8)$$

where

$$x_i = (x_i \bullet n, x_i \bullet pt), \qquad (9)$$

is information regarding the $i^{th}$ set of objects with

$$x_i \bullet n \subseteq \{1, ..., N_{C_a}\}, \qquad (10)$$

and

$$x_i \bullet pt \subseteq TY. \qquad (11)$$

Thus, we have

$$|\Theta_{I_a}| = (|TY| \cdot N_{C_a})^{|I_a|} \qquad (12)$$

The set of all intelligence reports $I_a$ in this representation becomes

$$J_a = \left\{\left\{m_i^{J_a}(\langle \theta_1, ..., \theta_{i-1}, x_i, \theta_{i+1}, ..., \theta_{|I_a|}\rangle)\right\}_{\{x_i\}}\right\}_{i=1}^{|I_a|} \qquad (13)$$

where

$$x_i \in \{(C_a^i \bullet n_j, C_a^i \bullet pt_j)\}_j \qquad (14)$$

is one of the propositions of intelligence report number $i$; $C_a^i$.

Let us begin the analysis of all intelligence by combining all mass functions $m_i^{J_a}$ of $J_a$, $\oplus J_a$.

Resulting from this combination we get

$$m_{\oplus J_a}(\langle x_1, x_2, ..., x_{|I_a|}\rangle) = \prod_{i=1}^{|I_a|} m_i^{J_a}(\langle \theta_1, ..., \theta_{i-1}, x_i, \theta_{i+1}, ..., \theta_{|I_a|}\rangle) \quad (15)$$

the basic probability number for each alternative hypothesis.

As we are also interested in the types of objects and their number regardless of their ordering, we sum up all contributions regarding the same type.

We let

$$m_{\oplus J_a}^{I_a}[(X \bullet n_j, X \bullet pt_j)] = \sum_{\substack{\langle x_1, x_2, ..., x_{|I_a|}\rangle \\ \left| X \bullet n_j = \bigoplus_{i|x_i \bullet pt = X \bullet pt_j} x_i \bullet n \right.}} m_{\oplus J_a}^{J_a}(\langle x_1, x_2, ..., x_{|I_a|}\rangle). \quad (16)$$

where $\bigoplus_i x_i \bullet n$ is the direct sum of all $x_i \bullet n$'s, not to be confused with Dempster's rule $\oplus$, as in $m_{\oplus J_a}^{I_a}$. The result is a set of elements, each element the sum of one element from every set $x_i \bullet n$, i.e.,

$$\bigoplus_i x_i \bullet n = \left\{ \sum_i y_i \middle| y_i \in x_i \bullet n \right\}, \quad (17)$$

e.g., the direct sum $\{1, 2\} \bigoplus \{2, 3\} = \{1+2, 1+3, 2+2, 2+3\} = \{3, 4, 4, 5\} = \{3, 4, 5\}$. This gives us information about different propositions in the initial representation with $\Theta_X$ as frame of discernment. The result of Eq. (16) will not be used in the selection process of finding a template with maximal fitness for the intelligence in $\chi_a$. However, it is a result in itself that may be communicated for other purposes.

## 2.1 An example

Let us observe an example with two intelligence reports and four possible vehicles. The first report has an uncertainty about whether the observation reported upon was of two main battle tanks (*MBT*) or two armored personnel carriers (*APC*), but with a strong preference for the first. It is initially represented as

$$m_1^{I_a}[(\{2\}, \{MBT\})] = 0.5,$$
$$m_1^{I_a}[(\{2\}, \{MBT, APC\})] = 0.3, \quad (18)$$
$$m_1^{I_a}(\Theta) = 0.2,$$

stating that we have a 0.5 basic probability in favor of two *MBTs*, a 0.3 basic probability in favor of two vehicles that are either *MBTs* or *APCs*.

The second report is uncertain both about the number of vehicles and the type of the vehicle observed;

$$m_2^{I_a}[(\{1, 2\}, \{MBT, APC\})] = 0.6, \quad (19)$$
$$m_2^{I_a}(\Theta) = 0.4,$$

stating that we have a 0.6 basic probability of one to two vehicles that are *MBTs* or *APCs*.

Representing these reports in the frame of Eq. (8), we obtain

$$m_1^{J_a}[\langle(\{2\}, \{MBT\}), \theta_2\rangle] = 0.5,$$
$$m_1^{J_a}[\langle(\{2\}, \{MBT, APC\}), \theta_2\rangle] = 0.3, \quad (20)$$
$$m_1^{J_a}[\langle\theta_1, \theta_2\rangle] = 0.2.$$

and

$$m_2^{J_a}[\langle\theta_1, (\{1, 2\}, \{MBT, APC\})\rangle] = 0.6, \quad (21)$$
$$m_2^{J_a}[\langle\theta_1, \theta_2\rangle] = 0.4.$$

Here, we have

$$J_a = \left\{ m_1^{J_a}, m_2^{J_a} \right\}. \quad (22)$$

We combine the two mass functions of $J_a$, Eqs. (20) and (21), to obtain

$$m_{\oplus J_a}(\langle(\{2\}, \{MBT\}), (\{1, 2\}, \{MBT, APC\})\rangle) = 0.3,$$
$$m_{\oplus J_a}(\langle(\{2\}, \{MBT\}), \theta_2\rangle) = 0.2,$$
$$m_{\oplus J_a}(\langle(\{2\}, \{MBT, APC\}), (\{1, 2\}, \{MBT, APC\})\rangle) = 0.18,$$
$$m_{\oplus J_a}(\langle(\{2\}, \{MBT, APC\}), \theta_2\rangle) = 0.12,$$
$$m_{\oplus J_a}(\langle\theta_1, (\{1, 2\}, \{MBT, APC\})\rangle) = 0.12,$$
$$m_{\oplus J_a}(\langle\theta_1, \theta_2\rangle) = 0.08. \quad (23)$$

using Eq. (15). This result will be used in the next section to select a template with maximum fitness towards the intelligence.

Temporarily, we return to the previous representation using $\Theta_X$ as the frame of discernment in order to obtain a basic probability assignment for each supported subset of all types of objects *TY*.

We sum up the contribution of Eq. (23) using Eq. (16), to receive

$$m_{\oplus J_a}^{I_a}[(\{2\}, \{MBT\})] = 0.3 + 0.2 = 0.5,$$
$$m_{\oplus J_a}^{I_a}[(\{0\}, \{MBT\})] = 0.18 + 0.12 + 0.12 + 0.08$$
$$= 0.5,$$
$$m_{\oplus J_a}^{I_a}[(\{3, 4\}, \{MBT, APC\})] = 0.18,$$
$$m_{\oplus J_a}^{I_a}[(\{2\}, \{MBT, APC\})] = 0.12,$$
$$m_{\oplus J_a}^{I_a}[(\{1, 2\}, \{MBT, APC\})] = 0.3 + 0.12 = 0.42,$$
$$m_{\oplus J_a}^{I_a}[(\{0\}, \{MBT, APC\})] = 0.2 + 0.08 = 0.28.$$

$$(24)$$

This summarizes the support for each supported subset of all types *TY*. Note, that the first two sum to 1.0, and the four last sum to 1.0 as these are two different assignments.

## 3   Templates

Comparing templates having specific propositions that are certain in what they are requesting with intelligence propositions that are not only uncertain but may also be nonspecific in what they are supporting can be a difficult task. The idea we use to handle this problem is to compare a candidate template with intelligence from the perspective of each and every subset of all possible types of objects *TY*.

In doing this we investigate how much support a subset of *TY* receives both directly and indirectly from intelligence and template, respectively. The support for a subset of *TY* is summed up from all propositions that are equal to or itself a subset of this subset of *TY*. This is similar to the calculation of belief from basic probability numbers in Dempster-Shafer theory, except that we are not summing up basic probability numbers but natural numbers representing the number of objects of the proposed types.

For example, from the perspective of {*MBT*, *APC*} a template proposition of "four *MBTs*" lend indirect support to {*MBT*, *APC*} since {*MBT*} is a subset of {*MBT*, *APC*}, and intelligence proposing "two *MBTs* or *APCs*" lend direct support to the subset. With the summed up numbers being four and two, respectively, we have a mismatch between template and intelligence from the perspective of {*MBT*, *APC*}. We use this method to rank all templates based on a fitness measure of template to intelligence matching taking all subsets of *TY* into account.

By using the result obtained by Eq. (15) from the combination of all mass functions in $J_a$, we compare different templates in order to find a template with maximum fitness towards the set of intelligence reports.

Let *T* be a set of all available templates $\{T_i\}$. Each template is represented by any number of slots $S_i^j$ where $S_i^j \cdot pt \in TY$ is a possible type from the set *TY* and $S_i^j \cdot n$ is the number of that type i $T_i$.

Based on the combination of all intelligence reports, Eq. (15), we evaluate all templates of $\{T_i\}$.

As we have several different alternative propositions in the intelligence regarding the type of objects and their corresponding number of objects, we need to compare each potential template with these alternatives and let each proposition influence the evaluation. For each template we find a measure of fitness between the template and each proposition in the intelligence, separately,

$$\pi_{\langle x_1, x_2, ..., x_{|I_a|} \rangle}(T_i). \tag{25}$$

We then make a linear combination where each measure of fitness is weighted by the basic probability number of that proposition,

$$m_{\oplus J_a}(\langle x_1, x_2, ..., x_{|I_a|} \rangle). \tag{26}$$

We get

$$\pi_{\oplus J_a}(T_i) = \sum_{\langle x_1, x_2, ..., x_{|I_a|} \rangle} m_{\oplus J_a}(\langle x_1, x_2, ..., x_{|I_a|} \rangle) \pi_{\langle x_1, x_2, ..., x_{|I_a|} \rangle}(T_i) \tag{27}$$

as the measure of fitness of $T_i$ towards all intelligence in $\chi_a$. This is the measure by which we rank all templates and make our selection of template.

In [6] we evaluated all templates $T_i$ by comparing each template against a set of intelligence reports with a single certain and specific proposition. This is here extended to handle intelligence reports with multiple uncertain and nonspecific propositions. We have

$$\pi_{\langle x_1, x_2, ..., x_{|I_a|} \rangle}(T_i) = \frac{1}{2}[\pi^1_{\langle x_1, x_2, ..., x_{|I_a|} \rangle}(T_i) + \pi^2_{\langle x_1, x_2, ..., x_{|I_a|} \rangle}(T_i)] \tag{28}$$

as a measure of fitness for template $T_i$ towards one of these multiple propositions $\langle x_1, x_2, ..., x_{|I_a|} \rangle$, where

$$\pi^1_{\langle x_1, x_2, ..., x_{|I_a|} \rangle}(T_i) = \min_j \left[ \pi^3_{\langle x_1, x_2, ..., x_{|I_a|} \rangle}(T_i | S_a^j \cdot pt) \right], \tag{29}$$

with $S_a^j \cdot pt \subseteq TY$, is a measure of fitness looking for a worst matching between $T_i$ and this proposition for all different subsets of all types *TY*.

Here,

$$\pi^3_{\langle x_1, x_2, ..., x_{|I_a|} \rangle}(T_i | S_a^j \cdot pt)$$

$$= \begin{cases} \max_{n \in SC_a(S_a^j \cdot pt)} \left\{ \min\left[ \frac{n}{ST_i(S_a^j \cdot pt)}, \frac{ST_i(S_a^j \cdot pt)}{n} \right] \right\}, ST_i(S_a^j \cdot pt) > 0 \\ 1 \qquad\qquad\qquad\qquad\qquad\qquad\qquad , ST_i(S_a^j \cdot pt) = 0 \end{cases}$$
$$\tag{30}$$

is a measure of fitness for template $T_i$ towards the same propositions $\langle x_1, x_2, ..., x_{|I_a|} \rangle$ from the perspective of $S_a^j \cdot pt$ only.

The second measure in Eq. (28),

$$\pi^2_{\langle x_1, x_2, ..., x_{|I_a|} \rangle}(T_i) = \max_{n \in SC_a(TY)} \left\{ \min\left[ \frac{n}{ST_i(TY)}, \frac{ST_i(TY)}{n} \right] \right\} \tag{31}$$

is only looking for the correct number of objects in $T_i$ and in proposition $\langle x_1, x_2, ..., x_{|I_a|} \rangle$ of the intelligence, regardless of object types.

While the first measure $\pi^1_{\langle x_1, x_2, ..., x_{|I_a|} \rangle}(T_i)$ measures the fitness of $T_i$ on a type-by-type basis demanding a perfect fit for all types to give a full score, the second measure ignores type entirely, and compare the number of objects of all intelligence in $\chi_a$ with the same in the template. While the first measure seems preferable it can be too extreme when considering missing data for same small number $S_i^j \bullet n$ of objects in $T_i$.

Note that $ST_i(S_a^j \bullet pt) \; \forall i, j$ in Eq. (30) and $ST_i(TY) \; \forall i$ in Eq. (31) can be precomputed using Eq. (32) below as they are independent of intelligence.

For each potential template $T_i$ we calculate the number of objects requested by the template from the perspective of subset $X \subseteq TY$ in Eq. (30), (31) as

$$\forall X \subseteq TY . ST_i(X) = \sum_{j | S_i^j \bullet pt \subseteq X \bullet pt} S_i^j \bullet n, \qquad (32)$$

and the number of objects supported by proposition $\langle x_1, x_2, ..., x_{|I_a|} \rangle$ of the intelligence from the perspective of subset $X \subseteq TY$ in Eq. (30), (31) as

$$\forall X \subseteq TY . SC_a(X | \langle x_1, x_2, ..., x_{|I_a|} \rangle) = \sum_{\substack{i | x_i \in \langle x_1, x_2, ..., x_{|I_a|} \rangle \\ x_i \bullet pt \subseteq X \bullet pt}} x_i \bullet n \qquad (33)$$

where $\bigoplus_i x_i \bullet n$ is the same direct sum of integer sets as in Eq. (17); each element in the resulting set the addition of one element from every set $x_i \bullet n$.

Here, we assume that $x_i \bullet n = \{0, ..., N_{C_a}\}$ and not $\{0\}$ when $x_i = \theta_i$. There is after all a difference between having a report that is uncertain about its proposition and being sure there is no object at all. The latter being a very strong statement.

While the fitness measure $\pi_{\oplus J_a}(\cdot)$ is used for aggregation from the current hierarchical level, we also need the basic probability of the highest ranked template for any further aggregation from the next hierarchical level.

We combine the intelligence, Eq. (15), with a basic probability assignment stating that the set of all templates is true,

$$m_T(\{T_i\}) = 1. \qquad (34)$$

Each focal element in the resulting combination support a subset of all templates. Through a fitness weighted transformation, these templates will share this support in relation to their fitness towards the corresponding focal element in the intelligence.

We find the basic probability number of a template $T_i$ as

$$m_{\oplus J_a}(T_i) = \sum_{\langle x_1, x_2, ..., x_{|I_a|} \rangle \supseteq T_i} \left\{ m_{\oplus J_a}(\langle x_1, x_2, ..., x_{|I_a|} \rangle) \right. \\ \left. \times \frac{\pi_{\langle x_1, x_2, ..., x_{|I_a|} \rangle}(T_i)}{\sum_{\langle x_1, x_2, ..., x_{|I_a|} \rangle \supseteq T_j} \pi_{\langle x_1, x_2, ..., x_{|I_a|} \rangle}(T_j)} \right\} \qquad (35)$$

using Eqs. (15) and (28).

## 3.1 An example continued

Let us evaluate templates based on a comparison with the result of Eq. (15) (in Eq. (23)). Let us assume we have two templates one with four main battle tanks (*MBTs*) and one with three armored personnel carriers (*APCs*). Our frame of discernment is $\Theta = \{MBT, APC\}$.

Table 1: Number allowed by templates ($ST_1$, $ST_2$) and supported by intelligence ($SC_a$) for different propositions $S_a^j \bullet pt$.

|  |  | {MBT} | {APC} | {MBT, APC} |
|---|---|---|---|---|
| $ST_1(\cdot)$ | $(\{4\}, \{MBT\})$ | 4 | 0 | 4 |
| $ST_2(\cdot)$ | $(\{3\}, \{APC\})$ | 0 | 3 | 3 |
| $SC_a(\cdot | \langle \cdot \rangle)$ | $\langle (\{2\}, \{MBT\}), (\{1,2\}, \{MBT, APC\}) \rangle$ | {2} | {0} | {3, 4} |
|  | $\langle (\{2\}, \{MBT\}), \theta_2 \rangle$ | {2} | {0} | $\{2, ..., N_{C_a}\}$ |
|  | $\langle (\{2\}, \{MBT, APC\}), (\{1,2\}, \{MBT, APC\}) \rangle$ | {0} | {0} | {3, 4} |
|  | $\langle (\{2\}, \{MBT, APC\}), \theta_2 \rangle$ | {0} | {0} | $\{2, ..., N_{C_a}\}$ |
|  | $\langle \theta_1, (\{1,2\}, \{MBT, APC\}) \rangle$ | {0} | {0} | $\{1, ..., N_{C_a}\}$ |
|  | $\langle \theta_1, \theta_2 \rangle$ | {0} | {0} | $\{0, ..., N_{C_a}\}$ |

Table 2: A measure of fitness $\pi^3_{\langle \cdot \rangle}(T_i | \cdot)$ between each template proposition and every proposition in fused intelligence.

|  | $T_1$ | | | $T_2$ | | |
|---|---|---|---|---|---|---|
|  | {MBT} | {APC} | {MBT, APC} | {MBT} | {APC} | {MBT, APC} |
| $\langle (\{2\}, \{MBT\}), (\{1,2\}, \{MBT, APC\}) \rangle$ | 2/4 | 1 | 4/4 | 1 | 0/3 | 3/3 |
| $\langle (\{2\}, \{MBT\}), \theta_2 \rangle$ | 2/4 | 1 | 4/4 | 1 | 0/3 | 3/3 |
| $\langle (\{2\}, \{MBT, APC\}), (\{1,2\}, \{MBT, APC\}) \rangle$ | 0/4 | 1 | 4/4 | 1 | 0/3 | 3/3 |
| $\langle (\{2\}, \{MBT, APC\}), \theta_2 \rangle$ | 0/4 | 1 | 4/4 | 1 | 0/3 | 3/3 |
| $\langle \theta_1, (\{1,2\}, \{MBT, APC\}) \rangle$ | 0/4 | 1 | 4/4 | 1 | 0/3 | 3/3 |
| $\langle \theta_1, \theta_2 \rangle$ | 0/4 | 1 | 4/4 | 1 | 0/3 | 3/3 |

We have $T = \{T_1, T_2\}$ where

$$S_1^1 \bullet pt = MBT, \qquad S_1^1 \bullet n = 4,$$

and

$$S_2^1 \bullet pt = APC, \qquad S_2^1 \bullet n = 3.$$

We also have fused intelligence according to Eq. (23) in Sect. 2.1.

In order to evaluate both templates $T_1$ and $T_2$ we calculate $\pi_{\oplus J_a}(T_1)$ and $\pi_{\oplus J_a}(T_2)$ to find their measure of fitness towards the fused intelligence. To calculate these measures we must first calculate the number of objects requested by the templates for each subset of $TY$ using Eq. (32). Using Eq. (33) with the fused intelligence give us the number of objects supported by different propositions in the fused intelligence. In Table 1 we have tabulated the result of $ST_1(\cdot)$ and $ST_2(\cdot)$ for the two templates and $SC_a(\cdot)$ for all propositions in the fused intelligence.

From the result in Table 1 we find $\pi_{\langle \cdot \rangle}^3(T_i|\cdot)$ the measure of fitness of each template towards every $\langle x_1, x_2, ..., x_{|I_a|} \rangle$ proposition in the intelligence given each subset of $TY$, in Table 2 using Eq. (30). In the same manner we find $\pi_{\langle \cdot \rangle}^2(\cdot)$, i.e., the same measure of fitness towards all $\langle x_1, x_2, ..., x_{|I_a|} \rangle$ propositions in the intelligence, but given the set of all types $TY$ (Table 3, using Eq. (31)).

Table 3: $\pi_{\langle \cdot \rangle}^2(T_i)$.

|  | $T_1$ | $T_2$ |
|---|---|---|
| $\langle (\{2\}, \{MBT\}), (\{1,2\}, \{MBT, APC\}) \rangle$ | 4/4 | 3/3 |
| $\langle (\{2\}, \{MBT\}), \theta_2 \rangle$ | 4/4 | 3/3 |
| $\langle (\{2\}, \{MBT, APC\}), (\{1,2\}, \{MBT, APC\}) \rangle$ | 4/4 | 3/3 |
| $\langle (\{2\}, \{MBT, APC\}), \theta_2 \rangle$ | 4/4 | 3/3 |
| $\langle \theta_1, (\{1,2\}, \{MBT, APC\}) \rangle$ | 4/4 | 3/3 |
| $\langle \theta_1, \theta_2 \rangle$ | 4/4 | 3/3 |

From Table 2 we find the minimum fitness $\pi_{\langle \cdot \rangle}^1(T_i)$ for both templates towards every proposition in the fused intelligence given all subsets of $TY$ (tabulated in Table 4, using Eq. (29)).

Table 4: $\pi_{\langle \cdot \rangle}^1(T_i)$.

|  | $T_1$ | $T_2$ |
|---|---|---|
| $\langle (\{2\}, \{MBT\}), (\{1,2\}, \{MBT, APC\}) \rangle$ | 2/4 | 0/3 |
| $\langle (\{2\}, \{MBT\}), \theta_2 \rangle$ | 2/4 | 0/3 |
| $\langle (\{2\}, \{MBT, APC\}), (\{1,2\}, \{MBT, APC\}) \rangle$ | 0/4 | 0/3 |
| $\langle (\{2\}, \{MBT, APC\}), \theta_2 \rangle$ | 0/4 | 0/3 |
| $\langle \theta_1, (\{1,2\}, \{MBT, APC\}) \rangle$ | 0/4 | 0/3 |
| $\langle \theta_1, \theta_2 \rangle$ | 0/4 | 0/3 |

Finally, we use Eq. (28) to take ½ of $\pi_{\langle \cdot \rangle}^1(\cdot|T_i)$ and $\pi_{\langle \cdot \rangle}^2(\cdot|T_i)$ for all propositions in Table 3 and Table 4 to get the results in Table 5, $\pi_{\langle \cdot \rangle}(T_i)$.

Table 5: $\pi_{\langle \cdot \rangle}(T_i)$.

|  | $T_1$ | $T_2$ |
|---|---|---|
| $\langle (\{2\}, \{MBT\}), (\{1,2\}, \{MBT, APC\}) \rangle$ | 3/4 | 1/2 |
| $\langle (\{2\}, \{MBT\}), \theta_2 \rangle$ | 3/4 | 1/2 |
| $\langle (\{2\}, \{MBT, APC\}), (\{1,2\}, \{MBT, APC\}) \rangle$ | 1/2 | 1/2 |
| $\langle (\{2\}, \{MBT, APC\}), \theta_2 \rangle$ | 1/2 | 1/2 |
| $\langle \theta_1, (\{1,2\}, \{MBT, APC\}) \rangle$ | 1/2 | 1/2 |
| $\langle \theta_1, \theta_2 \rangle$ | 1/2 | 1/2 |

Finally, using Eq. (27) to find a linear combination of the measure of fitness in Table 5 (from Eq. (28)) and the basic probability numbers of all propositions in Eq. (23), we obtain measures of fitness for both templates;

$$\pi_{\oplus J_a}(T_1) = 0.3\frac{3}{4} + 0.2\frac{3}{4} + 0.18\frac{1}{2} + 0.12\frac{1}{2} + 0.12\frac{1}{2} + 0.08\frac{1}{2}$$
$$= 0.625$$

and

$$\pi_{\oplus J_a}(T_2) = 0.3\frac{1}{2} + 0.2\frac{1}{2} + 0.18\frac{1}{2} + 0.12\frac{1}{2} + 0.12\frac{1}{2} + 0.08\frac{1}{2}$$
$$= 0.50.$$

As $\pi_{\oplus J_a}(T_1) > \pi_{\oplus J_a}(T_2)$, template $T_1$ is the preferred classification of the intelligence in cluster $\chi_a$.

Using Eq. (35) we find the basic probability of $T_1$ by combining Eq. (23) with $m_T(\{T_1, T_2\}) = 1$. We get

$$m_{\oplus J_a}(T_1) = 0.3\frac{\frac{3}{4}}{\frac{3}{4}} + 0.2\frac{\frac{3}{4}}{\frac{3}{4}} + 0.18\frac{\frac{1}{2}}{\frac{1}{2}+\frac{1}{2}} + 0.12\frac{\frac{1}{2}}{\frac{1}{2}+\frac{1}{2}}$$
$$+ 0.12\frac{\frac{1}{2}}{\frac{1}{2}+\frac{1}{2}} + 0.08\frac{\frac{1}{2}}{\frac{1}{2}+\frac{1}{2}} = 0.75$$

as the first two focal elements intersects to $\{T_1\}$ and the remainder to $\{T_1, T_2\}$.

# 4 An algorithm for evidential force aggregation

Summarizing the results of Sects. 2 and 3, we find an algorithm for force aggregation from evidential data as follows:

First, combine all intelligence for $\chi_a$ as represented in $J_a$. Secondly, calculate the basic probability number $m_{\oplus J_a}(\langle x_1, x_2, ..., x_{|I_a|} \rangle)$ for all propositions in the result from $\oplus J_a$, using Eq. (15).

At the same time, calculate the number of objects supported by intelligence $SC_a(\cdot|\langle \cdot \rangle)$ for each subset of all types $TY$ and for each proposition in the intelligence using Eq. (33), and calculate the number of objects requested by each template $ST_i(\cdot)$ for each subset of all types $TY$ and for each proposition in the intelligence, using Eq. (32).

From $SC_a(\cdot|\langle \cdot \rangle)$ and $ST_i(\cdot)$ we may calculate $\pi_{\langle x_1, x_2, ..., x_{|I_a|} \rangle}(T_i)$ for each template and each proposition in the intelligence by using Eqs. (28), (29), (30), (31).

Finally, calculate a measure of fitness for all templates $\pi_{\oplus J_a}(T_i)$ $\forall i$ based as a linear combination of all $\pi_{\langle x_1, x_2, ..., x_{|I_a|} \rangle}(T_i)$ using $m_{\oplus J_a}(\langle x_1, x_2, ..., x_{|I_a|} \rangle)$ and Eq. (27).

The unit that is aggregated from intelligence is $T_i$ for which $\pi_{\oplus J_a}(T_i)$ is maximal, Figure 2.

Finally, we calculate the support of $T_i$ using Eq. (35).

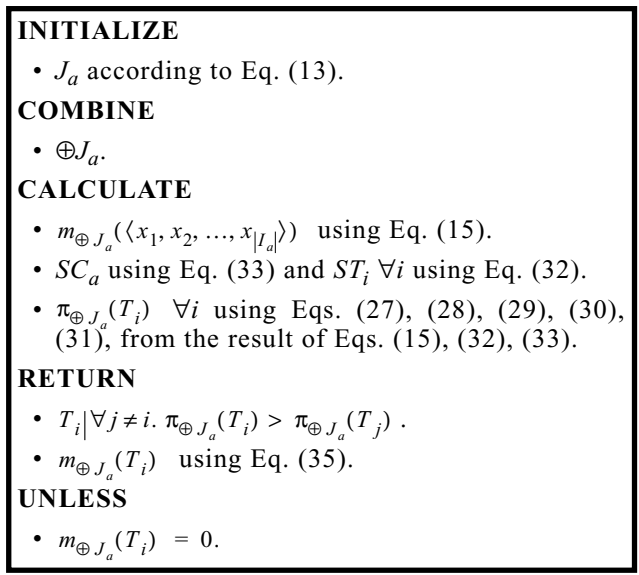

**INITIALIZE**
- $J_a$ according to Eq. (13).

**COMBINE**
- $\oplus J_a$.

**CALCULATE**
- $m_{\oplus J_a}(\langle x_1, x_2, ..., x_{|I_a|} \rangle)$ using Eq. (15).
- $SC_a$ using Eq. (33) and $ST_i \; \forall i$ using Eq. (32).
- $\pi_{\oplus J_a}(T_i) \; \forall i$ using Eqs. (27), (28), (29), (30), (31), from the result of Eqs. (15), (32), (33).

**RETURN**
- $T_i | \forall j \neq i. \; \pi_{\oplus J_a}(T_i) > \pi_{\oplus J_a}(T_j)$.
- $m_{\oplus J_a}(T_i)$ using Eq. (35).

**UNLESS**
- $m_{\oplus J_a}(T_i) = 0$.

Figure 2: An evidential force aggregation algorithm.

## 5 Conclusions

The evidential force aggregation method presented makes it possible to aggregate uncertain intelligence reports with multiple uncertain and nonspecific propositions into recognized forces using templates.

This is an extension in two ways compared to earlier methods [6]: (*i*) it handles intelligence reports that are statistically uncertain, (*ii*) it handles any number of such propositions. These propositions may continue to be specific or nonspecific in the sense that a proposition may support any subset of all possible object or unit types. With this extension we are able to aggregate general intelligence into units.

## References

[1] Johan Schubert, *On nonspecific evidence,* International Journal of Intelligent Systems, Vol. 8, No. 6, pp. 711–725, Jul. 1993.

[2] Mats Bengtsson, and Johan Schubert, *Dempster-Shafer clustering using potts spin mean field theory,* Soft Computing, Vol. 5, No. 3, pp. 215–228, June 2001.

[3] Johan Schubert, *Managing inconsistent intelligence,* Proceedings of the Third International Conference on Information Fusion, Paris, France, 10–13 Jul. 2000, pp. TuB4/10–16.

[4] Johan Schubert, *Robust Report Level Cluster-to-Track Fusion,* Proceedings of the Fifth International Conference on Information Fusion, Annapolis, USA, 8–11 Jul. 2002, pp. 913–918.

[5] Johan Schubert, *Clustering belief functions based on attracting and conflicting metalevel evidence,* Proceedings of the Ninth International Conference on Information Processing and Management of Uncertainty in Knowledge-based Systems, Annecy, France, 1–5 Jul. 2002, pp. 571–578, also in Intelligent Systems for Information Processing: From Representation to Applications, B. Bouchon-Meunier, L. Foulloy, R.R. Yager (Eds.), Elsevier Science Publishers, 2003, Amsterdam, to appear.

[6] Johan Schubert, *Reliable force aggregation using a refined evidence specification from Dempster-Shafer clustering,* Proceedings of the Fourth International Conference on Information Fusion, Montréal, Canada, 7–10 Aug. 2001, pp. TuB3/15–22.

[7] Joachim Biermann, *HADES - A knowledge-based system for message interpretation and situation determination,* in Tasks and methods in Applied Artificial Intelligence, Proceedings of the Eleventh International Conference on Industrial and Engineering Applications of Artificial Intelligence and Expert Systems, A. Pasqual del Pobil, J. Mira, M. Ali (Eds.), Castellón, Spain, 1–4 June 1998, pp. 707–716, Springer-Verlag (LNCS 1416), Berlin, 1998.

[8] Frank P. Lorenz, and Joachim Biermann, *Knowledge-based fusion of formets: discussion of an example,* Proceedings of the Fifth International Conference on Information Fusion, Annapolis, MD, USA, 8–11 Jul. 2002, pp. 374–379.

[9] Jason K. Johnson, and Ronald D. Chaney, *Recursive composition inference for force aggregation,* Proceedings of the Second International Conference on Information Fusion, Las Vegas, USA, 6–8 Jul. 1999, pp. 1187–1195.

[10] John Cantwell, Johan Schubert, and Johan Walter, *Conflict-based force aggregation,* Cd Proceedings of the Sixth International Command and Control Research and Technology Symposium, Annapolis, USA, 19–21 June 2001, Track 7, Paper 031, pp. 1–15.

[11] Arthur P. Dempster, *A generalization of Bayesian inference,* J. R. Statist. Soc. B, Vol. 30, No. 2, pp. 205–247, 1968.

[12] Glenn Shafer, *A mathematical theory of evidence,* Princeton University Press, Princeton, NJ, 1976.

[13] Glenn Shafer, *Perspectives on the theory and practice of belief functions,* International Journal of Approximate Reasoning, Vol. 4, No. 5/6, pp. 323–362, Sept.–Nov. 1990.

[14] Philippe Smets, and Robert Kennes, *The transferable belief model,* Artificial Intelligence, Vol. 66, No. 2. pp. 191–234, Mar. 1994.

[15] Philippe Smets, *Practical uses of belief functions,* Proceedings of the Fifteenth Conference on Uncertainty in Artificial Intelligence, Stockholm, Sweden, 30 Jul.–1 Aug. 1999, pp. 612–621.

[16] Philippe Smets, *Data fusion in the transferable belief model,* Proceedings of the Third International Conference on Information Fusion, Paris, France, 10–13 Jul. 2000, pp. PS/20–33.